\documentclass[sigconf]{aamas} 
\usepackage{balance}

\usepackage{balance} % for balancing columns on the final page

\setcopyright{none}
\renewcommand\footnotetextcopyrightpermission[1]{}
\acmSubmissionID{356}

\title[Language Models for Non-Cooperative Gamess]{Reasoning, Memorization, and Fine-Tuning Language Models for Non-Cooperative Games}

\author{Yunhao Yang}
\affiliation{
  \institution{The University of Texas at Austin}
  \city{Austin, Texas}
  \country{United States}}
\email{yunhaoyang234@utexas.edu}

\author{Leonard Berthellemy}
\affiliation{
  \institution{The University of Texas at Austin}
  \city{Austin, Texas}
  \country{United States}}
\email{leo.berth@utexas.edu}

\author{Ufuk Topcu}
\affiliation{
  \institution{The University of Texas at Austin}
  \city{Austin, Texas}
  \country{United States}}
\email{utopcu@utexas.edu}

\begin{abstract}
We develop a method that integrates the tree of thoughts and multi-agent framework to enhance the capability of pre-trained language models in solving complex, unfamiliar games. The method decomposes game-solving into four incremental tasks---game summarization, area selection, action extraction, and action validation---each assigned to a specific language-model agent. By constructing a tree of thoughts, the method simulates reasoning paths and allows agents to collaboratively distill game representations and tactics, mitigating the limitations of language models in reasoning and long-term memorization. Additionally, an automated fine-tuning process further optimizes the agents’ performance by ranking query-response pairs based on game outcomes, e.g., winning or losing. We apply the method to a non-cooperative game and demonstrate a 65 percent winning rate against benchmark algorithms, with an additional 10 percent improvement after fine-tuning. In contrast to existing deep learning algorithms for game solving that require millions of training samples, the proposed method consumes approximately 1000 training samples, highlighting its efficiency and scalability.
\end{abstract}

\keywords{Non-Cooperative Game, Language Model, Tree of Thoughts, Fine-Tuning}

\newcommand{\BibTeX}{\rm B\kern-.05em{\sc i\kern-.025em b}\kern-.08em\TeX}

\usepackage{amsmath,amsfonts}
\usepackage{algorithmic}
\usepackage{graphicx}
\usepackage{textcomp}
\usepackage{xcolor}
\usepackage{wrapfig}
\usepackage{svg}
\usepackage{listings}
\usepackage{multirow}
\usepackage{pgfplots}
\usepackage{wrapfig}
\pgfplotsset{compat=1.18} 

\definecolor{lightlightgray}{gray}{0.9}
\begin{document}

%%% The following commands remove the headers in your paper. For final 
%%% papers, these will be inserted during the pagination process.

\pagestyle{fancy}
\fancyhead{}

%%% The next command prints the information defined in the preamble.

\maketitle 

\section{Introduction}

While existing deep learning algorithms have demonstrated remarkable efficacy in tacking complex games such as chess and Go, these algorithms are computationally expensive. Existing algorithms typically rely on deep neural networks and reinforcement learning techniques to master these board games' intricate strategies and nuanced gameplay and require massive human-labeled data \cite{lapan2018deep, alphago, alphazero}.

The emergence of pre-trained language models obviates the necessity of training deep learning models for game playing, but their constraints on knowledge \cite{hu2024survey}, long-term memorization \cite{mem1, mem2, mem3, mem4}, and reasoning make them inadequate for playing games beyond their knowledge domain \cite{Kuo2023LargeLM, Toshniwal2021ChessAA, Yang2023OnTP}. Although several works address constraints \cite{Yang2022AutomatonBasedRO, huang2022language, davison2019commonsense, petroni2019language, edelkamp2002symbolic} through constructing abstractions or representations of the language outputs, they are inadequate or exceedingly inefficient in the phase of encapsulating the exponential growth of the size of game states.

We develop a method that integrates \emph{tree of thoughts} \cite{yao2024tree} and \emph{multi-agent framework} \cite{chen2024smurfs, qiao2024autoact} to distill game representations and tactics from language models. 
The method decomposes the distillation process into four smaller, incremental tasks, with four language models designated as agents, each responsible for a specific task. By building a tree of thoughts, the method simulates branching paths of reasoning and the agents follow these paths to exchange information by formulating queries to other agents. In particular, two agents distill current game representation, which includes the knowledge of the game state, winning conditions, and current objectives. They pass the representation to the third agent and the third agent extracts a tactic based on the representation. Finally, the fourth agent evaluates whether the tactic aligns with the current objectives and broadcasts the feedback.

The proposed method addresses the limitations of language models in handling complex, unfamiliar games. By constructing a tree of thoughts, the method eliminates the need for reasoning through the entire game. Furthermore, the multi-agent collaboration for representation distillation enables each language model to retain task-specific information with linear memory growth, as opposed to exponential expansion.

In addition, we develop an automated fine-tuning process to refine the language models based on feedback from game-solving. We deploy two players, both utilizing the proposed method, to play the game and independently collect query-response pairs from the four agents. We assign higher ranks to the query-response pairs from the winning player. Subsequently, we fine-tune each language model agent separately using the ranked pairs specific to its task, allowing for more effective adaptation and improvement of their task-solving capabilities.

We apply the proposed method to a newly defined non-cooperative game and demonstrate a 65 percent winning rate against benchmark algorithms utilizing language models. Moreover, the fine-tuning process results in an additional 10 percent increase in the winning rate over all the benchmarks. In contrast to existing deep learning algorithms for game-solving, which typically require millions of training samples, our approach achieves this 10 percent improvement with approximately 1000 samples, highlighting its efficiency and scalability.

\section{Related Works}

Existing deep learning methods, such as AlphaGo \cite{alphago}, AlphaZero \cite{alphazero} and DeepChess \cite{david2016deepchess} have showcased the ability to surpass human experts through extensive self-play and neural network training. However, these algorithms require extensive training and large computational power.

Existing approaches \cite{abdelnabi2023llm, duan2024gtbench, topsakal2024benchmarking, feng2024chessgpt} ask language models to decide the tactics in a game. They have shown the capability of language models in playing games such as Tik-Tac-Toe and chess. However, they rely on the language model containing prior knowledge of the game, hence inadequate to newly designed games beyond the language model's knowledge.

Furthermore, the constraints inherent in language models pertaining to long-term memorization and reasoning significantly circumscribe their utility within intricate gaming contexts \cite{mem1, mem2, mem3, mem4}. The inability for (and the impossibility of) memorizing the game history significantly reduces the language model's performance in playing games \cite{Kuo2023LargeLM, Toshniwal2021ChessAA, Yang2023OnTP}. 

In contrast to existing approaches, the method we propose leverages the tree of thoughts framework and a multi-agent collaboration system to address the limitations of language models in reasoning about unseen games while also reducing memory requirements. This method decomposes the game-solving process into multiple tasks and requires each language model to focus solely on its task, thereby reducing cognitive load and improving overall efficiency in reasoning and decision-making. Additionally, the method allows each model to maintain only task-relevant information, which significantly reduces memory overhead.

\section{Preliminary}

\paragraph{Non-Cooperative Game} The \emph{extensive form} of a non-cooperative game a tuple $\mathcal{G} = (P, Q, F, I, A, T, U)$, where
\begin{itemize}
\item $P = (1, ..., p)$ is a set of $p$ rational \emph{Players}.
\item $Q = \{q_1, ..., q_k\}$ is a set of \emph{Game States}.
\item $F = \{ q_{f_1}, ..., q_{f_d} \} \subset Q$ is a set of \emph{Termination States}.
\item $I: P \times Q \rightarrow Q$ is a function that maps a player and a game state to another game state. It indicates the game state from the player's perspective. In full-information games, the function maps each state to itself. In partial-information games, it may map a state to other states due to the player's misinformation.
\item $A = \{a_1,..., a_m\}$ is a set of \emph{Actions} that a player can take.
\item $T: Q\times P\times A \rightarrow Q$ is a transition function. It transits from one state to another based on which player is taking which action.
\item $U: F \rightarrow \mathbb{R}^p$ is a \emph{Utility} function that returns a list of $p$ numerical values indicating the utility of $p$ players at the termination state.
\end{itemize}

\paragraph{Goedendag} A variant of the game Stratego that is under development. 
It is a two-player board game where each player commands an army of pieces, each with a specific rank, and the objective is to capture the opponent’s flag (a special piece).
We present the game rules in Section \ref{sec: method} and more detailed descriptions of this game in the supplementary.

\begin{figure}[t]
    \centering
    \includegraphics[width=0.9\linewidth]{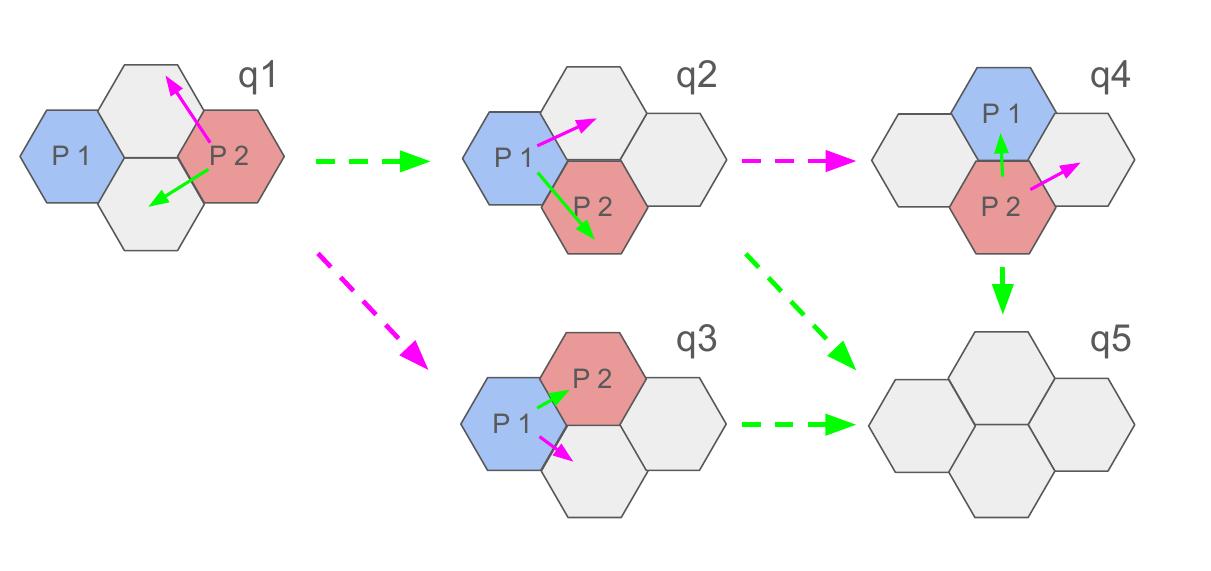}
    \caption{An example of the Goedendag game: The example shows two players $p1, p2$ and five states $q1,...,q5$. The game board comprises four hexagons; the blue and red hexagons indicate where $p1$ and $p2$'s pieces are located. The solid arrows are actions the player takes, and the dashed arrows are state transitions. The action with the same color triggers each transition. $q5$ is a terminated state of drawing, neither of them wins.}
    \Description[Goedendag]{Goedendag Example}
    \label{fig: game-example}
\end{figure}

It is a non-cooperative game $\mathcal{G} = (P, Q, F, I, A, T, U)$ with the following components:
\begin{itemize}
    \item $P = (p1, p2)$, there are two players in the game.
    \item $Q = \{q1, ..., qk\}$, each state corresponds to a layout of a game board, including the location and rank of each piece.
    \item $F \subset Q$, each terminate state indicates a player wins or draws the game.
    \item $I: P \times Q \rightarrow Q$, each player does not know all the ranks of the opponent's pieces.
    \item $A = \{a_1,..., a_m\}$, each action indicates a player's move of a piece.
    \item $T: Q\times P\times A \rightarrow Q$.
    \item $U: F \rightarrow \{-1,0,1\}^p$, a player can win, lose, or draw.
\end{itemize}
An example of the Goedendag is in Figure \ref{fig: game-example}.

Consider a newly designed non-cooperative game, e.g., Goedendag, where the game developer provides a written description of the rules. The written description can be viewed as a user manual to instruct players how to play and win the game.

\paragraph{Written Description of the Rules}
A written description $\mathcal{R}$ of the rules of a game $\mathcal{G} = (P, Q, F, I, A, T, U)$ is a set of natural language sentences that describe a function $R: Q \times P \rightarrow pow(A)$, where $pow(A)$ is the power set (set of all subsets) of $A$. Intuitively, $\mathcal{R}$ describes the sets of allowable actions for each player at each state. We present an example of $\mathcal{R}$ in Section \ref{sec: method}.

% \paragraph{Tool library: }
% A library $\mathcal{L}$ is a set of tools for users to query information about the game or modify the game states. The tool library $\mathcal{L} = \{\mathcal{I}, \mathcal{T}, \mathcal{U}\}$ in our setting includes
% \begin{itemize}
%     \item a function $\mathcal{I}: P \rightarrow Q$ that returns the game state from a player's perspective,
%     \item a function $\mathcal{T}: P \times A \rightarrow Q$ that takes a transition to another game state based on the input action, this function automatically tracks the current game state,
%     \item a function $\mathcal{U}: P \rightarrow \mathbb{R}$ returns the utility of a player.
% \end{itemize}

\paragraph{Language Model for Game Solving}
A language model takes a text-based input prompt and outputs a text-based response. By formulating the input prompts, we can query the language model to generate tactics (actions), understand game states, or predict utilities. However, directly querying the language model to solve games requires the language model containing prior knowledge of the game, e.g., well-known games such as chess and tic-tac-toe. 

We aim to use language models to solve unseen games whose rules are beyond their knowledge domain. In this work, we choose Goedendag---a variant of an existing game that is not released publicly---as the unseen game and use language models to extract game tactics. At a high level, we propose a method that iteratively sends game-relevant information to language models and queries for tactics/actions until reaching a termination state.

The proposed method formulates \emph{task-specific prompts} to guide the language models such that the action $a$ returns from the language model satisfies:
(1) For a player $p \in P$ at a state $q \in Q$, $a \in R(q, p)$, i.e., the action complies with the game rule.
(2) Maximize the player's utility.

\begin{figure*}[t]
    \centering
    \includegraphics[width=\textwidth]{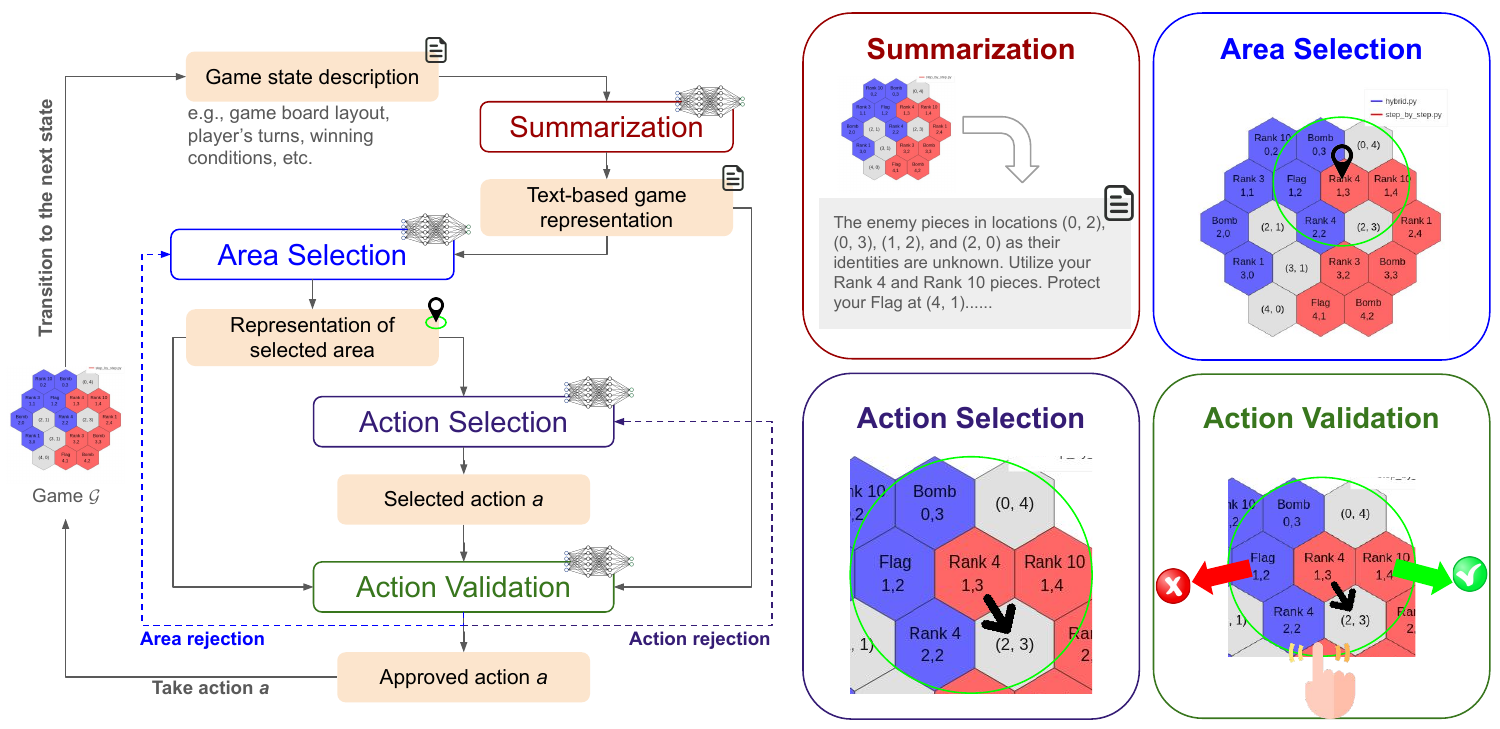}
    \caption{Demonstration of the method. The left figure shows the tree of thought that connects multiple tasks to solve the game. The right figure shows four tasks, where we assign a language model agent to solve each task.}
    \Description[Pipeline]{Overall Pipeline}
    \label{fig:ToT global presentation}
\end{figure*}

\section{Methodology}
\label{sec: method}

Consider a non-cooperative game $\mathcal{G} = (P, Q, F, I, A, T, U)$ with a written description of the game rules $\mathcal{R}$.
We design a method that enhances pre-trained language models' ability to solve complex, unfamiliar games by integrating the tree of thoughts framework with a multi-agent system. The method decomposes the game-solving process into four tasks: game summarization, area selection, action selection, and action validation. We assign each task to a separate pre-trained language model, denoted a \emph{language-model agent}, allowing for a collaborative and structured approach to game representation and decision-making.

The method employs the tree of thoughts to simulate multiple reasoning paths, enabling agents to extract game representations and tactics (actions). In particular, the summarization and area selection tasks aim to extract and estimate a game representation from the current state. The tree of thoughts passes the game representation to the agent for area selection. Then, the action selection task aims to extract a game tactic for the current player at the current state. The validation task checks whether the extracted tactic complies with the rules, provided by the written descriptions, and signals to other agents if the tactic fails to meet the rules. We demonstrate the method in Figure \ref{fig:ToT global presentation}.

In this section, we use the Goedendag as a running example to clarify the method.

\paragraph{Overview}
Given the game $\mathcal{G} = (P, Q, F, I, A, T, U)$, the method starts from a player $p \in P$ at a state $q \in Q$. The method guides four language-model agents working collaboratively to extract a game tactic, i.e., action, $a \in A$. If the action $a$ complies with the rules: $a \in R(q, p)$, the game will transit to a new state $q' = T(p, q, a)$.

If the new state $q'$ is one of the termination states: $q' \in F$, the method stops the game and computes the utility for the player: $U(q')_p$. Otherwise, the method extracts a new action given the new state and repeats this procedure until reaching a termination state.

\paragraph{Prompt Format}
Recall that we assign four language-model agents to four tasks and each language model $M_i$ takes the rule descriptions and a task-specific prompt as inputs, and returns text-based responses. Using the tree of thoughts to integrate the four agents, they collaboratively return an action $a \in A$ for the current player $p \in P$. We send the same rule description and different task-specific prompts to each language-model agent, in the following format:
\begin{lstlisting}[basicstyle=\small]
<Written Descriptions of Rules>.
You are player <player_number>.
<Task-Specific Prompt>.
\end{lstlisting}

The written descriptions of rules are global information included in the prompts to all the language-model agents, while the task-specific prompts are different for each agent.

The written description of rules $\mathcal{R}$ are text-based game manuals provided by the game designer.
In the Goedendag example, a written description of rules includes the description of the categories of pieces, the rules of moving the pieces, and the winning conditions:
\begin{lstlisting}[basicstyle=\small]
1. Players: player 0 and player 1. Each player will play turn by turn.
2. Types of pieces: Rank 1, ......, Rank 10, Bomb, and Flag.
3. Moving pieces: All pieces are movable except the bombs and the flag.
4. Attacking pieces: The piece moving into the tile is the Attacker and the piece that was already occupying that tile is the Defender......
5. Winning conditions: One of the player's pieces moves to the tile of the opponent's flag.
\end{lstlisting}

We attached the complete description in the supplementary.

A \emph{task-specific prompt} includes instructions to the language-model agent on what information we want to retrieve and the information retrieved from other agents. We will present more details on the task-specific prompt in Section \ref{sec: component}.

\textbf{Note: } The descriptions of rules are provided by the game designer. We use the Goedendag as an example for demonstration and empirical analysis, but they can be generalized to any non-cooperative games with provided descriptions of rules.

\paragraph{Error Handling}
As shown in Figure \ref{fig:ToT global presentation}, the last task before returning an action is action validation, where we use a language-model agent to check whether the selected action complies with the game rules. If the action fails to comply with the rules, we will loop back and re-query for a new action. We repeat this procedure until the action is approved, i.e., complies with the rules, or reaches the maximum number of repetitions defined by the user. We will present more details in Section \ref{sec: component} \emph{Action Validation}.

\subsection{Tasks}
\label{sec: component}
The method breaks the game-solving procedure into four tasks and assigns four language-model agents to these tasks. The method builds a tree of thoughts that formulates task-specific prompts to guide the agents in completing their tasks.

\paragraph{Summarization} 
The first task is using a language-model agent $M_1$ to extract a text-based game representation of the current game state. We denote this extraction procedure as \emph{summarization}.

Recall that every agent takes the rule description $\mathcal{R}$ and a task-specific prompt as inputs. We present the task-specific prompt for $M_1$ below.
\begin{lstlisting}[basicstyle=\small]
The current game state is:
<Game State Description>.

Summarize the current game state and how the game state could contribute to the objective.
\end{lstlisting}
The game state description is a text-based description of the current game state. In the Goedendag example, a game state description consists of the size of the game board, the positions of both players' pieces, and their categories:
\begin{lstlisting}[basicstyle=\small]
The board is composed of hexagonal tiles, each one having up to 6 neighbors. They are labeled with 2D coordinates. 
For a tile (i,j), its neighbors are the tiles (i+1,j), (i,j+1), (i-1,j), (i,j-1), (i-1,j+1), (i+1,j-1).

Description of the board: 
Tile (0, 2): Your Rank 3; Tile (0, 3): Your Rank 1;......
Tile (2, 4): Enemy unknown piece; Tile (3, 0): Enemy Rank 2;......
Tile (0, 4): Empty;......
\end{lstlisting}
$M_1$ returns a text-based game representation $\text{rep}(\mathcal{G})$, which encodes the current game state $q \in Q$, the current player $p$, a subset of allowable actions $A' \subseteq A$, and a set of all possible state transitions $\{T(q, p, a): a \in A'\}$. Note that the game state descriptions can be varied depending on the game state. In contrast, the text-based game representation is in a formalized, consistent structure across the entire game.
We present a sample game representation in the Goedendag game below.
\vspace{15pt}
\begin{lstlisting}[basicstyle=\small]
    Game representation returned from the language-model agent
@Your Pieces:
- Rank 3 at (0, 2): A strong piece that can engage in battle with lower ranks.
- Bomb at (2, 1): This piece is non-movable but can take down lower ranks if attacked.
- Flag at (2, 0): This is a critical piece; protecting it is paramount since losing it means losing the game.
......
Enemy Pieces:
- (1, 4), (2, 3), (2, 4): Unknown, posing potential threats, especially adjacent to your key pieces.
......@
\end{lstlisting}

This game representation encapsulates the relevant details of the current game state in a structured and coherent format, reducing the difficulty of game interpretation and facilitating the agents' reasoning and decision-making processes in the subsequent tasks. We empirically demonstrate the importance of this summarization task in Section \ref{sec: abalation}.

\paragraph{Area Selection}
Define \emph{area} $Q'$ as a subset of the game states: $Q' \subseteq Q$. Recall that the game states transit based on the current state, the current player, and the action: $T: Q \times P \times A \rightarrow Q$. Given the current game state $q \in Q$ and the current player $p \in P$, the area selection task aims to produce a subset of states $Q' \subseteq Q$ such that
\begin{equation}
\label{eq: subset-state}
    \forall_{q' \in Q'} \ \exists_{a \in A} \ T(q, p, a) = q'.
\end{equation}
By fixing the size of the area, i.e., the number of states in $Q'$, the method avoids the need to memorize the exponentially grown game states. Instead, the method only memorizes the selected area, which results in a linear growth in the memory.

In the area selection task, the method formulates a task-specific prompt that consists of the text-based game representation $\text{rep}(\mathcal{G})$ obtained from $M_1$ and an instruction on selecting a subset of game states.

A language-model agent $M_2$ takes the descriptions of rules $\mathcal{R}$ and the task-specific prompt as inputs. It returns a text-based representation $\text{rep}(\mathcal{G}')$ that encodes the information (e.g., states, player, actions, possible transitions) of $\mathcal{G}'$, where $\mathcal{G}' = (p, Q', F, I, A, T, U)$, $p \in P$ is the current player and $Q' \subseteq Q$ is the set of selected states satisfying Equation \ref{eq: subset-state}. We show an example of the representation with a selected area in the next blue text box.

\begin{figure}[t]
    \centering
    \includegraphics[width=0.8\linewidth]{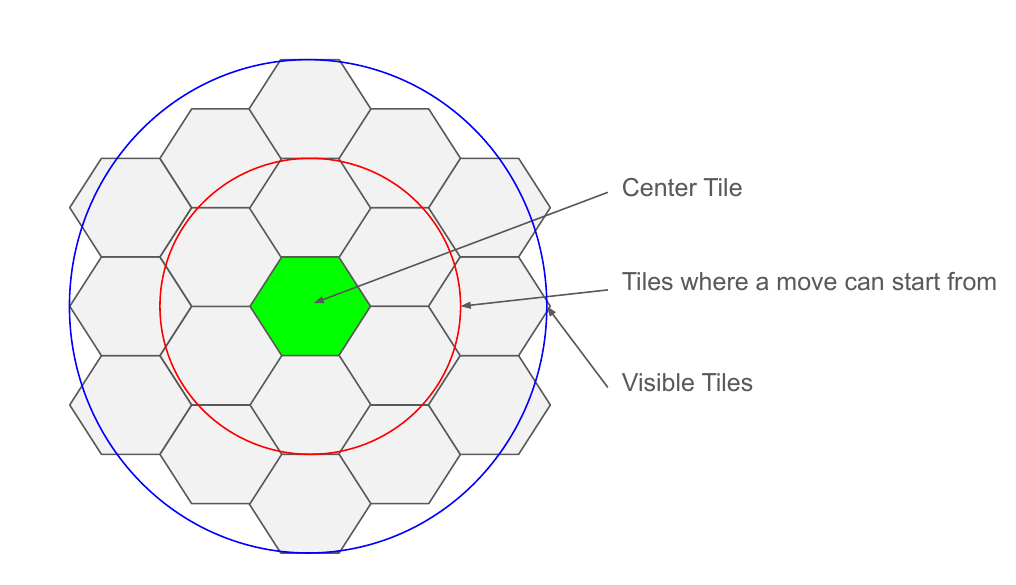}
    \Description[area example]{Example of a selected area}
    \caption{An example of a selected area.}
    \label{fig:area description}
    \vspace{-12pt}
\end{figure}

In the Goedendag game, game states correspond to the layouts of the game board. Hence, we can choose a subset of states by choosing a sub-area on the board. The task-specific prompt is:
\begin{lstlisting}[basicstyle=\small]
<Text-Based Game Representation>
Return the coordinates of the center tile of the area you want to select.
Return a description of the tile and its adjacent tiles, including the pieces on them and their owners.
\end{lstlisting}
The method first requests the language model to select a tile in the game board and considers it as the center tile of the selected area. Only pieces at the center tile or its adjacent tiles can move. And only pieces within a two-tile distance from the center tile are visible. Figure \ref{fig:area description} presents a demonstration of a selected area.

After selecting the area, $M_2$ returns a text-based representation of the game $\text{rep}(\mathcal{G})$ with only the selected area, i.e., the selected subset of states. We present an example below.
\begin{lstlisting}[basicstyle=\small]
            An area selected by a language-model agent
@My **Rank 10** piece at (1,2). (1,3): No piece present. 
My **Rank 3** piece at (0,2) has no available moves.
......@
\end{lstlisting}

\paragraph{Action Selection}
Given a game representation $\text{rep}(\mathcal{G}')$ for the selected area returned from $M_2$, where $\mathcal{G}' = (p, Q', F, I, A, T, U)$, the method then extracts an action $a$ such that $T(q, p, a) = q'$ and $q, q' \in Q'$.

In particular, a language-model agent $M_3$ takes the description of rules $\mathcal{R}$ and a task-specific prompt as inputs, then outputs an action $a$. The task-specific prompt consists of the representation $\text{rep}(\mathcal{G}')$ (e.g., the previous blue text box) and a short instruction ``choose an action from the selected area.''

In the Goedendag example, the method selects an action that moves a piece located within the red circle in Figure \ref{fig:area description}.
\begin{lstlisting}[basicstyle=\small]
            An action selected by the language-model agent
@Move the Rank 10 piece from (1, 2) to (1, 3).@
\end{lstlisting}

\paragraph{Action Validation}
Given the game representation $\text{rep}(\mathcal{G})$ from $M_1$ (summarization task), the representation $\text{rep}(\mathcal{G}')$ with the selected area from $M_2$, and the selected action $a$ from $M_3$, the method uses a language-model agent $M_4$ to determine whether the action complies with the rules described in $\mathcal{R}$.

The task-specific prompt for the action validation task consists of $\text{rep}(\mathcal{G})$, $\text{rep}(\mathcal{G}')$, $a$, and instruction ``Does this action comply with all the rules of the game?'' 

The agent $M_4$ takes the description of rules and the task-specific prompt as input, then returns a signal indicating whether the action complies with the rules. The agent returns one of the following three signals:

(1) If $M_4$ returns ``approval,'' the player can take this action and proceed to the next game state.

(2) If $M_4$ returns ``area rejection,'' indicating the selected area does not include any valid action, hence the method passes the rejection signal to $M_2$ and requests a new round of area selection. 

(3) The ``action rejection'' signal indicates valid actions exist in the selected area but the selected action violates the rules. If $M_4$ rejects the action, the method passes the rejection signal to $M_3$ and queries for a different action $a'$.

We present several sample outputs from $M_4$ in Goedendag:
\begin{lstlisting}[basicstyle=\small]
      Validation results returned from the language-model agent
@Approval: The Rank 10 piece can move safely to (1, 3) as it is unoccupied, allowing for future tactical positions.

Action rejection: (0, 2) is occupied by my piece.

Area rejection: there are no movable pieces in the selected area.@
\end{lstlisting}

The method proceeds the game if and only if $M_4$ returns ``approval.'' The approved action $a$ triggers a transition to a new game state $q' = T(q, p, a)$. If the new state is not a termination state, the method goes back to the first task ``summarization'' with a game state description of the new state.

\section{Experiment}

We apply the proposed method to play the Goedendag game against several benchmark algorithms. Recall that the Goedendag game requires two players. One player uses the proposed method to decide actions and the other uses the benchmark algorithm. We first evaluate the method by the two metrics described in Section \ref{sec: metric} and demonstrate how the proposed method outperforms the benchmarks.

Then, we perform a set of ablation studies to demonstrate the necessity of the four agents by removing one of the agents and showing performance degradation.

Lastly, we fine-tune the language models and show that the fine-tuned models further improve the proposed method.

\subsection{Experiment Setting}
\label{sec: metric}

\paragraph{Evaluation Metric} We use two metrics to evaluate the proposed method and other benchmarks.

\textit{(1) Win Rate: } Let $p_i$ and $p_j$ be the two players, $w_1$ be the number of games $p_1$ wins and $w_2$ be the number of games $p_2$ wins, the \emph{win rate} $r_w(p_i)$ of player $p_i$ is 
\begin{equation}
    \label{eq: win-rate}
    r_w(p_i) = \frac{w_1}{w_1 + w_2}.
\end{equation}
When we compute the win rates, we only count the games where a winning player exists.
If a player returns actions that do not comply with the rules, we randomly select an action that meets the rules for the player.

\textit{(2) Error Rate: } Let $\tilde A_i = \tilde a_1 \tilde a_2 \tilde a_3 ......$ be a sequence of actions player $p_i$ takes in the game, from the initial state to the termination state. Let $\varepsilon_i$ be the number of actions $\tilde a_j \in \tilde A_i$ that violate the rules described in $\mathcal{R}$, $|P|$ be the number of players, and $|\tilde A_i|$ be the total number of actions $p_i$ have taken to complete the game, the error rate of one game $\mathcal{E}$ is the number of actions violating the rules over the total number of actions:
\begin{equation}
\label{eq: error-rate}
    \mathcal{E} = \sum_{i=1}^{|P|} \frac{\varepsilon_i}{|\tilde A_i|} .
\end{equation}

If the proposed method or the benchmark returns actions that violate the rules, we replace these actions with randomly generated actions that comply with the rules and proceed to the next state, avoiding the game being interrupted by the rule violations.

\paragraph{Benchmark}
We select three benchmark methods to determine actions for a player. For a game $\mathcal{G} = (P, Q, F, I, A, T, U)$, all the benchmark methods return an action $a \in A$ for the player $p \in P$ based on the current state $q \in Q$.

\textit{(1) Direct Query: } This method directly sends the written descriptions of rules $\mathcal{R}$ and the game state description (defined in the \emph{summarization} task in Section \ref{sec: component}) into a language model and queries for an action:
\begin{lstlisting}[basicstyle=\small]
<Written Descriptions of Rules>.

The current game state is:
<Game State Description>.

What action should the current player take?
\end{lstlisting}
The language model returns a text-based description to a selected action $a \in A$, in the same format with $M_3$'s outputs in the \emph{action selection} task.

\textit{(2) Chain-of-Thought (CoT) Query: } This benchmark method builds an input prompt based on the chain-of-thought technique \cite{kojima2022large}. The prompt format is
\begin{lstlisting}[basicstyle=\small]
<Written Descriptions of Rules>.

The current game state is:
<Game State Description>.

Proceed by the following instructions:
<first instruction>
<second instruction>
...
\end{lstlisting}
Particular to the Goedendag game, we formulate the instructions as the following:
\begin{lstlisting}[basicstyle=\small]
1: Describe the board.
2: Describe all the allowable actions that comply with the rules.
3: Select the action that maximizes the possibility of winning.
4: Describe the selected action.
\end{lstlisting}
A language model takes the descriptions and instructions as inputs, then outputs an action $a \in A$ in natural language, e.g., ``move the Rank 1 piece from (0,0) to (1,0).''

The CoT query method uses a single zero-shot language model, as opposed to multiple language models in our method, hence it is less costly. We compare it against our method to show the performance-cost trade-off.
In contrast to the direct query method, the CoT query enables the language model to reason through the game before determining the action.

\textit{(3) Random: } This method first extracts all the valid actions at the current game state and then randomly selects an action $a$ from the set of valid actions. If there are $n$ valid actions, then the probability of each action being chosen is $1/n$.

\subsection{Benchmark Comparison}

\begin{figure}[t]
    \centering
    \includegraphics[height=0.45\linewidth]{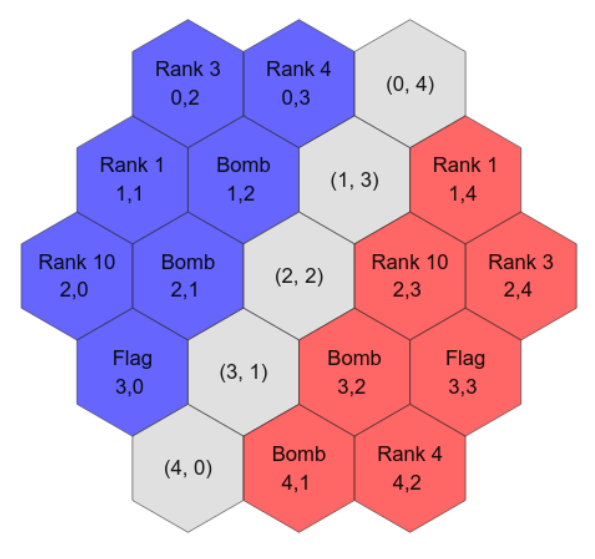}
    \includegraphics[height=0.45\linewidth]{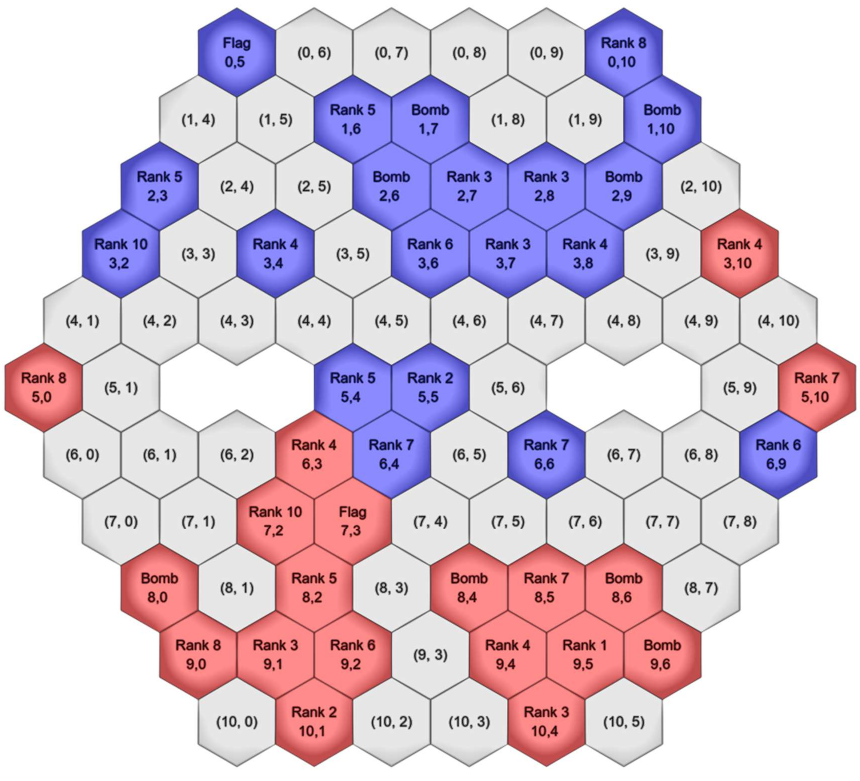}
    \caption{Example of a 4x4 board (left) and a 10x10 board (right).}
    \Description[board]{board example}
    \label{fig: board}
\end{figure}

We evaluate our proposed method against the three benchmarks over the win and error rates. In the experiments, we use \emph{GPT-4o-mini} in our method and benchmarks that require language models. We conduct pair-wise evaluations, where each evaluation consists of \textbf{5 sets} of games, and each set includes \textbf{40 games}. We split each set of 40 games into two game boards, 20 games over a 4x4 board and 20 games over a 10x10 board. We present the board layouts in Figure \ref{fig: board}. Note that the initial positions of pieces are randomly assigned based on the written descriptions of rules presented in the supplementary.

For each pair-wise evaluation, we compute the mean and standard deviation of our method's win rate against the benchmark. For example, if our method wins 15, 20, 25, 30, and 35 games in the five sets of games and loses the remaining games, the mean of its win rate is 25/40 and the standard deviation is $\sqrt{50}$. We report our method's win rates against all the benchmarks in Table \ref{tab:win rate}. The results indicate that our method achieves above 60 percent win rates against the benchmarks, significantly higher than the benchmarks' win rates (40 percent).

Then, we evaluate error rates for our method and the benchmarks for every game using Equation \ref{eq: error-rate} and present the average error rates across 200 games (5 sets x 40 games) in Table \ref{tab: error-rate}. Additionally, we record the average number of language-model tokens consumed in a game in Table \ref{tab: error-rate}, denoted as \emph{cost}.

The results showcase that our method achieves a significantly lower error rate than the benchmarks requiring language models (\emph{Random} does not query language model hence there is no error and cost). However, the downside of our proposed method is the cost, due to the use of four language-model agents and back-and-forth communications between the agents.

\begin{table}[t]
    \centering
    \begin{tabular}{||c|c|c||}
        \hline
        Benchmark & Win rate \\
        \hline
        Ours vs. Direct query & $0.647 \pm 0.225$ \\
        Ours vs. CoT query & $0.611 \pm 0.201$ \\
        Ours vs. Random & $0.662 \pm 0.136$ \\
        \hline
        Ours w/o summary vs. CoT query & $0.586 \pm 0.205$ \\
        Ours w/o summary vs. Random & $0.618 \pm 0.217$ \\
        \hline
        Ours w/o area vs. CoT query & $0.615 \pm 0.231$ \\
        Ours w/o area vs. Random & $0.633 \pm 0.159$ \\
        \hline
    \end{tabular}
    \vspace{4pt}
    \caption{Win rates against benchmarks (mean $\pm$ standard deviation).}
    \label{tab:win rate}
    \vspace{-15pt}
\end{table}

\begin{table}[t]
    \centering
    \begin{tabular}{||c|c|c||}
        \hline
        Benchmark & Error rate & Cost \\
        \hline
        Ours & $0.0575$ & $26211$ \\
        Ours w/o area & $0.313$ & $23519$ \\
        Ours w/o summary & $0.0417$ & $17881$ \\
        Direct query & $0.210$ & $7481$ \\
        CoT query & $0.362$ & $6976$ \\
        Random & $N/A$ & $N/A$ \\
        \hline
    \end{tabular}
    \vspace{4pt}
    \caption{Comparison between the error rate and cost (the number of language-model tokens consumed in one game).}
    \label{tab: error-rate}
    \vspace{-22pt}
\end{table}

\subsection{Ablation Study}
\label{sec: abalation}
We perform a set of ablation studies to indicate the necessity of the tasks in our proposed method. Recall that the method consists of four tasks: summarization, area selection, action selection, and validation. The former two tasks extract game representations and the latter two tasks decide actions. While the tasks for deciding actions are mandatory, we remove the summarization and area selection tasks and observe how these two tasks impact the performance.

We revise the tree of thoughts in the proposed method and formulate two variances of our method, denoted as \emph{ours w/o area} and \emph{ours w/o summary}. 

\noindent
- Ours w/o summary skips the summarization task and directly passes the game state description into the area selection task. The rest of the tasks proceed as normal.

\noindent
- Ours w/o area skips the area selection task. The method passes the text-based game representation directly to the action selection task and removes the backward communication from action validation to area selection.

We evaluate the two variances of the method against several benchmarks and present the results in Table \ref{tab:win rate} and \ref{tab: error-rate}. From Table \ref{tab:win rate}, we observe that removing the summarization task decreases the win rate by approximately 5 percent while removing the area selection task decreases the win rate by 2 percent. From the perspective of winning the game, the summarization task distills a complete game representation that helps the understanding of the current game state, leading to a significant improvement in the win rate.

On the other hand, Table \ref{tab: error-rate} shows that our method without area selection results in a 5x higher error rate compared to our method with area selection. The area selection task distills a game representation for a subset of states (i.e., a sub-area of a board), significantly lowering the number of actions to choose from. Hence, the language-model agent for the action selection task can easily determine an action that complies with the rules from a much smaller set of actions.
Therefore, we have demonstrated the necessity of both tasks in our method.

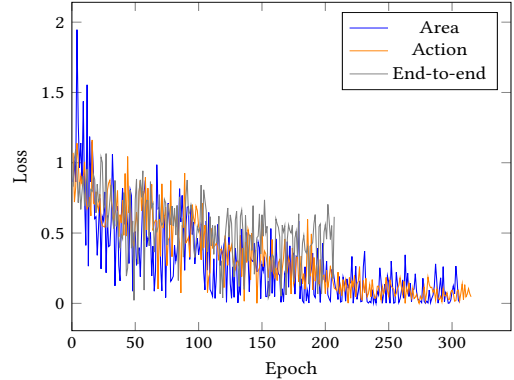
\begin{figure}[t]
    \centering
    \resizebox{0.8\linewidth}{0.6\linewidth}{\begin{tikzpicture}
\begin{axis}[
ylabel=Loss,
xlabel=Epoch,
xmin = 0,
legend pos=north east,
]
\addplot[blue] table [x=step, y=train_loss, col sep=comma] {figures/csv/area_result.csv};
\addlegendentry{Area}

\addplot[orange] table [x=step, y=train_loss, col sep=comma] {figures/csv/selection_result.csv};
\addlegendentry{Action}

\addplot[gray] table [x=step, y=train_loss, col sep=comma] {figures/csv/joint_result.csv};
\addlegendentry{End-to-end}
\end{axis}
\end{tikzpicture}}
    \caption{Cross entropy loss at every epoch during the fine-tuning procedure. The area and action agents converge to a lower loss compared to the end-to-end agent, indicating a potential better performance.}
    \Description[loss]{fine-tuning loss}
    \label{fig: loss}
    % \vspace{-12pt}
\end{figure}

\subsection{Fine-Tuning}

To further refine our approach, we implement a self-play strategy, where our method competes against itself to gather input-output pairs from the four tasks performed by the winning agent. These collected pairs are subsequently used to fine-tune the language-model agents. Due to the limitation of the computing resources, we only collect data and fine-tune the agents for area selection and action selection tasks. In all the fine-tuning procedures, we use the default supervised fine-tuning algorithm provided by OpenAI \cite{liu2023gpt} with early stopping (at convergence) \cite{dodge2020fine} to fine-tune the GPT-40-mini we used in our experiments.

We collect 1060 input/output pairs from each of the two tasks (area and action selection) across 100 games from the 4x4 board. Then, we fine-tune three language-model agents:
\begin{itemize}
    \item Area agent: a language model fine-tuned by the data collected from the area selection task.
    \item Action agent: a language model fine-tuned by the data collected from the action selection task.
    \item End-to-end agent: a language model fine-tuned by all the collected data.
\end{itemize}
We present the fine-tuning losses for the three agents in Figure \ref{fig: loss}.

Subsequently, we substitute the original agents in our method with the fine-tuned agents in various configurations to evaluate how the fine-tuning influences the win rates. We assess four replacement strategies: 
\begin{enumerate}
    \item replacing only the original agent for the area selection task with the fine-tuned area agent, 
    \item replacing only the agent for the action selection task with the fine-tuned action agent, 
    \item replacing agents for both tasks with their respective fine-tuned agents, 
    \item replacing agents for both tasks with the end-to-end fine-tuned agent.
\end{enumerate}

\begin{figure*}[t]
    \centering
    % \includesvg[width=0.4\linewidth]{figures/plots/win-rate-mas-5x5.svg}
    % \includesvg[width=0.4\linewidth]{figures/plots/win-rate-mas-10x10.svg}
    % \includesvg[width=0.4\linewidth]{figures/plots/win-rate-cot-5x5.svg}
    % \includesvg[width=0.4\linewidth]{figures/plots/win-rate-cot-10x10.svg}
    % \includesvg[width=0.4\linewidth]{figures/plots/win-rate-direct-5x5.svg}
    % \includesvg[width=0.4\linewidth]{figures/plots/win-rate-direct-10x10.svg}
    \includegraphics[width=0.38\linewidth]{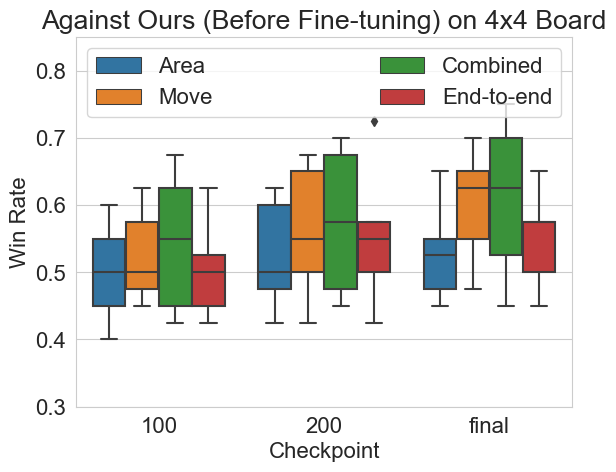}
    \includegraphics[width=0.38\linewidth]{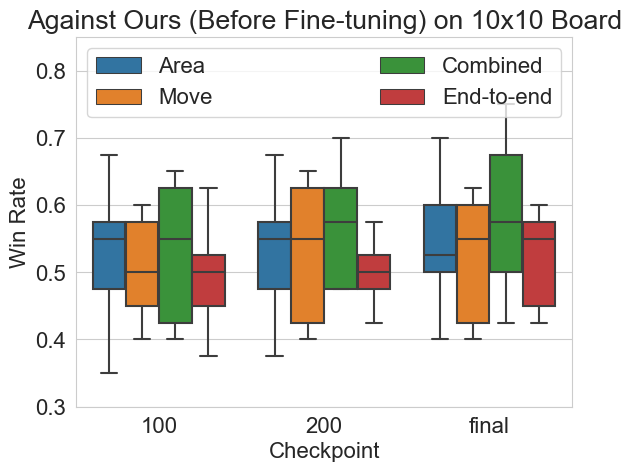}
    \includegraphics[width=0.38\linewidth]{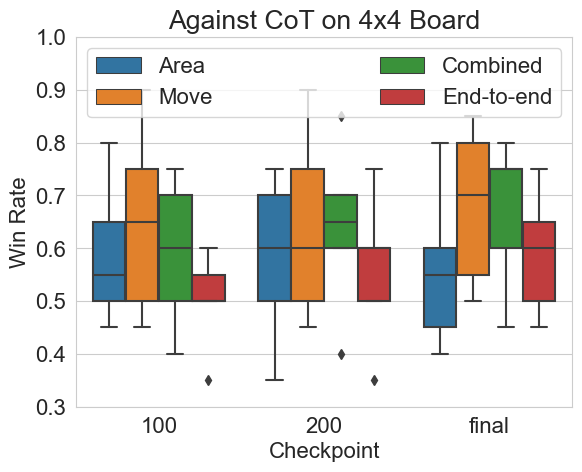}
    \includegraphics[width=0.38\linewidth]{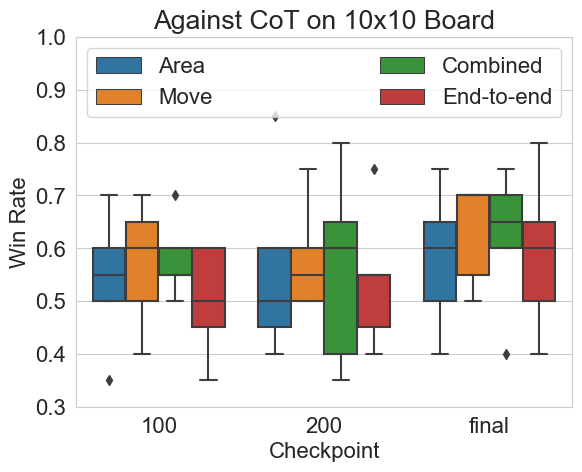}
    \includegraphics[width=0.38\linewidth]{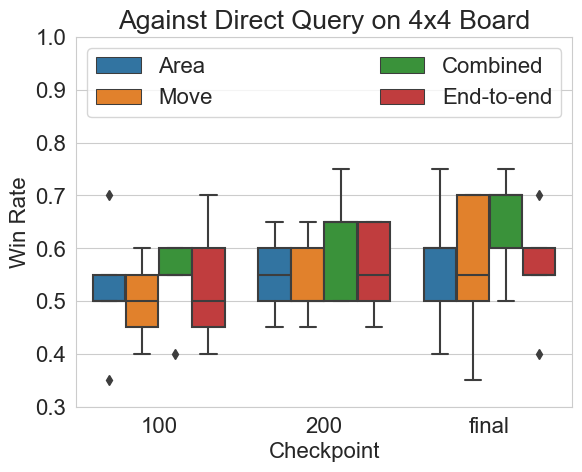}
    \includegraphics[width=0.38\linewidth]{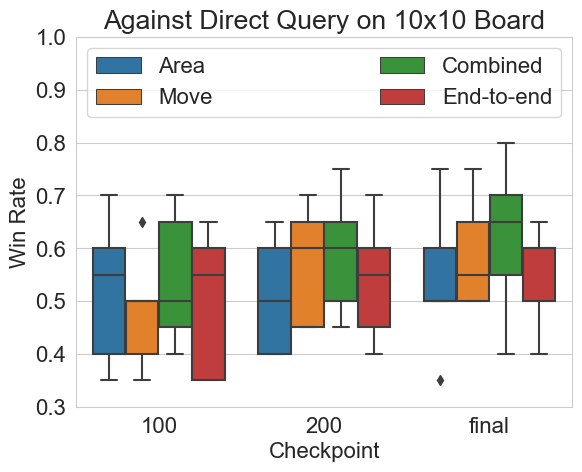}
    \caption{Win rate of our method with fine-tuned language-model agents against the benchmarks. 
    We record a checkpoint for each language model at every 100 training epochs. In the figure, "Area,'' ``move,'' ``combined,'' and ``end-to-end'' refer to replacement strategies (1), (2), (3), and (4), respectively.
    Our results indicate that independently fine-tuning agents for distinct tasks results in the highest win rate across all benchmarks, demonstrating the effectiveness of the proposed multi-agent approach.}
    \Description[fine-tune]{fine-tune}
    \label{fig: fine-tune-win-rate}
    % \vspace{-12pt}
\end{figure*}

Figure \ref{fig: fine-tune-win-rate} shows the win rate of our method, employed with the four strategies, against the benchmarks and our method before refinement. We fine-tune the agents over 4x4 boards and validate them over 10x10 boards. We show an average of 10 percent improvement over the win rates, demonstrating the effectiveness of our fine-tuning procedure. Additionally, fine-tuning each agent independently achieves significantly higher win rates compared with using a single end-to-end model for multiple tasks, showcasing the necessity of our multi-agent approach.
\section{Conclusion}

We develop a method that integrates the tree of thoughts with a multi-agent framework, effectively enhancing pre-trained language models' ability to solve complex and unfamiliar games. The method alleviates the limitation of language models on long-term memorization and reasoning by breaking down the game-solving process into incremental tasks. The results, including a 65 percent winning rate with an additional 10 percent improvement post-fine-tuning, demonstrate the method's potential, especially given its efficiency in requiring only a fraction of the training samples compared to traditional deep learning approaches.

For future research, this method can be extended to a wider variety of game types, such as cooperative games. Additionally, constructing a memory bank to retain past game states could further enhance the method's long-term memorization capabilities. Another avenue for improvement lies in refining the fine-tuning process by implementing more advanced ranking mechanisms based on detailed game metrics, which may result in further performance enhancements.

\bibliographystyle{ACM-Reference-Format} 
\balance
\bibliography{references}


\begin{thebibliography}{27}

%%% the Web address.

\ifx \showCODEN    \undefined \def \showCODEN     #1{\unskip}     \fi
\ifx \showDOI      \undefined \def \showDOI       #1{#1}\fi
\ifx \showISBNx    \undefined \def \showISBNx     #1{\unskip}     \fi
\ifx \showISBNxiii \undefined \def \showISBNxiii  #1{\unskip}     \fi
\ifx \showISSN     \undefined \def \showISSN      #1{\unskip}     \fi
\ifx \showLCCN     \undefined \def \showLCCN      #1{\unskip}     \fi
\ifx \shownote     \undefined \def \shownote      #1{#1}          \fi
\ifx \showarticletitle \undefined \def \showarticletitle #1{#1}   \fi
\ifx \showURL      \undefined \def \showURL       {\relax}        \fi
% invisible to TeX
\providecommand\bibfield[2]{#2}
\providecommand\bibinfo[2]{#2}
\providecommand\natexlab[1]{#1}
\providecommand\showeprint[2][]{arXiv:#2}

\bibitem[\protect\citeauthoryear{Abdelnabi, Gomaa, Sivaprasad, Sch{\"o}nherr, and Fritz}{Abdelnabi et~al\mbox{.}}{2023}]%
        {abdelnabi2023llm}
\bibfield{author}{\bibinfo{person}{Sahar Abdelnabi}, \bibinfo{person}{Amr Gomaa}, \bibinfo{person}{Sarath Sivaprasad}, \bibinfo{person}{Lea Sch{\"o}nherr}, {and} \bibinfo{person}{Mario Fritz}.} \bibinfo{year}{2023}\natexlab{}.
\newblock \bibinfo{title}{Llm-deliberation: Evaluating LLMs with interactive multi-agent negotiation games}.
\newblock
\newblock


\bibitem[\protect\citeauthoryear{Biderman, Prashanth, Sutawika, Schoelkopf, Anthony, Purohit, and Raf}{Biderman et~al\mbox{.}}{2023}]%
        {mem4}
\bibfield{author}{\bibinfo{person}{Stella Biderman}, \bibinfo{person}{USVSN~Sai Prashanth}, \bibinfo{person}{Lintang Sutawika}, \bibinfo{person}{Hailey Schoelkopf}, \bibinfo{person}{Quentin~G. Anthony}, \bibinfo{person}{Shivanshu Purohit}, {and} \bibinfo{person}{Edward Raf}.} \bibinfo{year}{2023}\natexlab{}.
\newblock \bibinfo{title}{Emergent and Predictable Memorization in Large Language Models}.
\newblock
\newblock


\bibitem[\protect\citeauthoryear{Chen, Liang, and Wang}{Chen et~al\mbox{.}}{2024}]%
        {chen2024smurfs}
\bibfield{author}{\bibinfo{person}{Junzhi Chen}, \bibinfo{person}{Juhao Liang}, {and} \bibinfo{person}{Benyou Wang}.} \bibinfo{year}{2024}\natexlab{}.
\newblock \bibinfo{title}{Smurfs: Leveraging Multiple Proficiency Agents with Context-Efficiency for Tool Planning}.
\newblock
\newblock


\bibitem[\protect\citeauthoryear{David, Netanyahu, and Wolf}{David et~al\mbox{.}}{2016}]%
        {david2016deepchess}
\bibfield{author}{\bibinfo{person}{Omid~E David}, \bibinfo{person}{Nathan~S Netanyahu}, {and} \bibinfo{person}{Lior Wolf}.} \bibinfo{year}{2016}\natexlab{}.
\newblock \showarticletitle{Deepchess: End-to-end deep neural network for automatic learning in chess}. In \bibinfo{booktitle}{\emph{Artificial Neural Networks and Machine Learning--ICANN 2016: 25th International Conference on Artificial Neural Networks, Barcelona, Spain, September 6-9, 2016, Proceedings, Part II 25}}. \bibinfo{publisher}{Springer}, \bibinfo{address}{New York, USA}, \bibinfo{pages}{88--96}.
\newblock


\bibitem[\protect\citeauthoryear{Davison, Feldman, and Rush}{Davison et~al\mbox{.}}{2019}]%
        {davison2019commonsense}
\bibfield{author}{\bibinfo{person}{Joe Davison}, \bibinfo{person}{Joshua Feldman}, {and} \bibinfo{person}{Alexander~M. Rush}.} \bibinfo{year}{2019}\natexlab{}.
\newblock \showarticletitle{Commonsense Knowledge Mining from Pretrained Models}. In \bibinfo{booktitle}{\emph{Proceedings of the 2019 Conference on Empirical Methods in Natural Language Processing and the 9th International Joint Conference on Natural Language Processing, {EMNLP-IJCNLP}}}, \bibfield{editor}{\bibinfo{person}{Kentaro Inui}, \bibinfo{person}{Jing Jiang}, \bibinfo{person}{Vincent Ng}, {and} \bibinfo{person}{Xiaojun Wan}} (Eds.). \bibinfo{publisher}{Association for Computational Linguistics}, \bibinfo{address}{Hong Kong, China}, \bibinfo{pages}{1173--1178}.
\newblock
\urldef\tempurl%
\url{https://doi.org/10.18653/V1/D19-1109}
\showDOI{\tempurl}


\bibitem[\protect\citeauthoryear{de~Wynter, Wang, Sokolov, Gu, and Chen}{de~Wynter et~al\mbox{.}}{2023}]%
        {mem2}
\bibfield{author}{\bibinfo{person}{Adrian de Wynter}, \bibinfo{person}{Xun Wang}, \bibinfo{person}{Alex Sokolov}, \bibinfo{person}{Qilong Gu}, {and} \bibinfo{person}{Si-Qing Chen}.} \bibinfo{year}{2023}\natexlab{}.
\newblock \bibinfo{title}{An Evaluation on Large Language Model Outputs: Discourse and Memorization}.
\newblock
\newblock


\bibitem[\protect\citeauthoryear{Dodge, Ilharco, Schwartz, Farhadi, Hajishirzi, and Smith}{Dodge et~al\mbox{.}}{2020}]%
        {dodge2020fine}
\bibfield{author}{\bibinfo{person}{Jesse Dodge}, \bibinfo{person}{Gabriel Ilharco}, \bibinfo{person}{Roy Schwartz}, \bibinfo{person}{Ali Farhadi}, \bibinfo{person}{Hannaneh Hajishirzi}, {and} \bibinfo{person}{Noah Smith}.} \bibinfo{year}{2020}\natexlab{}.
\newblock \bibinfo{title}{Fine-tuning pretrained language models: Weight initializations, data orders, and early stopping}.
\newblock
\newblock


\bibitem[\protect\citeauthoryear{Duan, Zhang, Diffenderfer, Kailkhura, Sun, Stengel-Eskin, Bansal, Chen, and Xu}{Duan et~al\mbox{.}}{2024}]%
        {duan2024gtbench}
\bibfield{author}{\bibinfo{person}{Jinhao Duan}, \bibinfo{person}{Renming Zhang}, \bibinfo{person}{James Diffenderfer}, \bibinfo{person}{Bhavya Kailkhura}, \bibinfo{person}{Lichao Sun}, \bibinfo{person}{Elias Stengel-Eskin}, \bibinfo{person}{Mohit Bansal}, \bibinfo{person}{Tianlong Chen}, {and} \bibinfo{person}{Kaidi Xu}.} \bibinfo{year}{2024}\natexlab{}.
\newblock \bibinfo{title}{Gtbench: Uncovering the strategic reasoning limitations of LLMs via game-theoretic evaluations}.
\newblock
\newblock


\bibitem[\protect\citeauthoryear{Duong and Solomon}{Duong and Solomon}{2023}]%
        {mem3}
\bibfield{author}{\bibinfo{person}{D. Duong} {and} \bibinfo{person}{B.~D. Solomon}.} \bibinfo{year}{2023}\natexlab{}.
\newblock \bibinfo{title}{Analysis of large-language model versus human performance for genetics questions}.
\newblock
\newblock


\bibitem[\protect\citeauthoryear{Edelkamp}{Edelkamp}{2002}]%
        {edelkamp2002symbolic}
\bibfield{author}{\bibinfo{person}{Stefan Edelkamp}.} \bibinfo{year}{2002}\natexlab{}.
\newblock \showarticletitle{Symbolic exploration in two-player games: Preliminary results}. In \bibinfo{booktitle}{\emph{The International Conference on AI Planning \& Scheduling (AIPS), Workshop on Model Checking}}. \bibinfo{pages}{40--48}.
\newblock


\bibitem[\protect\citeauthoryear{Feng, Luo, Wang, Tang, Yang, Shao, Mguni, Du, and Wang}{Feng et~al\mbox{.}}{2023}]%
        {feng2024chessgpt}
\bibfield{author}{\bibinfo{person}{Xidong Feng}, \bibinfo{person}{Yicheng Luo}, \bibinfo{person}{Ziyan Wang}, \bibinfo{person}{Hongrui Tang}, \bibinfo{person}{Mengyue Yang}, \bibinfo{person}{Kun Shao}, \bibinfo{person}{David Mguni}, \bibinfo{person}{Yali Du}, {and} \bibinfo{person}{Jun Wang}.} \bibinfo{year}{2023}\natexlab{}.
\newblock \showarticletitle{ChessGPT: Bridging Policy Learning and Language Modeling}. In \bibinfo{booktitle}{\emph{Advances in Neural Information Processing Systems}}, \bibfield{editor}{\bibinfo{person}{Alice Oh}, \bibinfo{person}{Tristan Naumann}, \bibinfo{person}{Amir Globerson}, \bibinfo{person}{Kate Saenko}, \bibinfo{person}{Moritz Hardt}, {and} \bibinfo{person}{Sergey Levine}} (Eds.). \bibinfo{publisher}{NIPS}, \bibinfo{address}{New Orleans, LA, USA}.
\newblock


\bibitem[\protect\citeauthoryear{Hu, Huang, Ilhan, Tekin, Liu, Kompella, and Liu}{Hu et~al\mbox{.}}{2024}]%
        {hu2024survey}
\bibfield{author}{\bibinfo{person}{Sihao Hu}, \bibinfo{person}{Tiansheng Huang}, \bibinfo{person}{Fatih Ilhan}, \bibinfo{person}{Selim Tekin}, \bibinfo{person}{Gaowen Liu}, \bibinfo{person}{Ramana Kompella}, {and} \bibinfo{person}{Ling Liu}.} \bibinfo{year}{2024}\natexlab{}.
\newblock \bibinfo{title}{A survey on large language model-based game agents}.
\newblock
\newblock


\bibitem[\protect\citeauthoryear{Huang, Abbeel, Pathak, and Mordatch}{Huang et~al\mbox{.}}{2022}]%
        {huang2022language}
\bibfield{author}{\bibinfo{person}{Wenlong Huang}, \bibinfo{person}{Pieter Abbeel}, \bibinfo{person}{Deepak Pathak}, {and} \bibinfo{person}{Igor Mordatch}.} \bibinfo{year}{2022}\natexlab{}.
\newblock \showarticletitle{Language Models as Zero-Shot Planners: Extracting Actionable Knowledge for Embodied Agents}. In \bibinfo{booktitle}{\emph{International Conference on Machine Learning}} \emph{(\bibinfo{series}{Proceedings of Machine Learning Research}, Vol.~\bibinfo{volume}{162})}. \bibinfo{publisher}{{PMLR}}, \bibinfo{address}{Baltimore, Maryland, {USA}}, \bibinfo{pages}{9118--9147}.
\newblock


\bibitem[\protect\citeauthoryear{Kojima, Gu, Reid, Matsuo, and Iwasawa}{Kojima et~al\mbox{.}}{2022}]%
        {kojima2022large}
\bibfield{author}{\bibinfo{person}{Takeshi Kojima}, \bibinfo{person}{Shixiang~Shane Gu}, \bibinfo{person}{Machel Reid}, \bibinfo{person}{Yutaka Matsuo}, {and} \bibinfo{person}{Yusuke Iwasawa}.} \bibinfo{year}{2022}\natexlab{}.
\newblock \showarticletitle{Large Language Models are Zero-Shot Reasoners}. In \bibinfo{booktitle}{\emph{Advances in Neural Information Processing Systems}}, \bibfield{editor}{\bibinfo{person}{Sanmi Koyejo}, \bibinfo{person}{S.~Mohamed}, \bibinfo{person}{A.~Agarwal}, \bibinfo{person}{Danielle Belgrave}, \bibinfo{person}{K.~Cho}, {and} \bibinfo{person}{A.~Oh}} (Eds.). \bibinfo{publisher}{NIPS}, \bibinfo{address}{New Orleans, LA, USA}.
\newblock


\bibitem[\protect\citeauthoryear{Kuo, Hsueh, and Tsai}{Kuo et~al\mbox{.}}{2023}]%
        {Kuo2023LargeLM}
\bibfield{author}{\bibinfo{person}{Mu-Tien Kuo}, \bibinfo{person}{Chih-Chung Hsueh}, {and} \bibinfo{person}{Richard Tzong-Han Tsai}.} \bibinfo{year}{2023}\natexlab{}.
\newblock \bibinfo{title}{Large Language Models on the Chessboard: A Study on ChatGPT's Formal Language Comprehension and Complex Reasoning Skills}.
\newblock
\newblock


\bibitem[\protect\citeauthoryear{Lapan}{Lapan}{2018}]%
        {lapan2018deep}
\bibfield{author}{\bibinfo{person}{Maxim Lapan}.} \bibinfo{year}{2018}\natexlab{}.
\newblock \bibinfo{booktitle}{\emph{Deep Reinforcement Learning Hands-On: Apply modern RL methods, with deep Q-networks, value iteration, policy gradients, TRPO, AlphaGo Zero and more}}.
\newblock \bibinfo{publisher}{Packt Publishing Ltd}, \bibinfo{address}{Birmingham, UK}.
\newblock


\bibitem[\protect\citeauthoryear{Liu, Zheng, Du, Ding, Qian, Yang, and Tang}{Liu et~al\mbox{.}}{2023}]%
        {liu2023gpt}
\bibfield{author}{\bibinfo{person}{Xiao Liu}, \bibinfo{person}{Yanan Zheng}, \bibinfo{person}{Zhengxiao Du}, \bibinfo{person}{Ming Ding}, \bibinfo{person}{Yujie Qian}, \bibinfo{person}{Zhilin Yang}, {and} \bibinfo{person}{Jie Tang}.} \bibinfo{year}{2023}\natexlab{}.
\newblock \showarticletitle{GPT understands, too}.
\newblock \bibinfo{journal}{\emph{AI Open}}  \bibinfo{volume}{1} (\bibinfo{year}{2023}), \bibinfo{pages}{0--11}.
\newblock


\bibitem[\protect\citeauthoryear{McGrath, Kapishnikov, Toma{\v{s}}ev, Pearce, Wattenberg, Hassabis, Kim, Paquet, and Kramnik}{McGrath et~al\mbox{.}}{2022}]%
        {alphazero}
\bibfield{author}{\bibinfo{person}{Thomas McGrath}, \bibinfo{person}{Andrei Kapishnikov}, \bibinfo{person}{Nenad Toma{\v{s}}ev}, \bibinfo{person}{Adam Pearce}, \bibinfo{person}{Martin Wattenberg}, \bibinfo{person}{Demis Hassabis}, \bibinfo{person}{Been Kim}, \bibinfo{person}{Ulrich Paquet}, {and} \bibinfo{person}{Vladimir Kramnik}.} \bibinfo{year}{2022}\natexlab{}.
\newblock \showarticletitle{Acquisition of chess knowledge in alphazero}.
\newblock \bibinfo{journal}{\emph{Proceedings of the National Academy of Sciences}} \bibinfo{volume}{119}, \bibinfo{number}{47} (\bibinfo{year}{2022}), \bibinfo{pages}{e2206625119}.
\newblock


\bibitem[\protect\citeauthoryear{Peng, Wang, and Deng}{Peng et~al\mbox{.}}{2023}]%
        {mem1}
\bibfield{author}{\bibinfo{person}{Zhencan Peng}, \bibinfo{person}{Zhizhi Wang}, {and} \bibinfo{person}{Dong Deng}.} \bibinfo{year}{2023}\natexlab{}.
\newblock \showarticletitle{Near-Duplicate Sequence Search at Scale for Large Language Model Memorization Evaluation}.
\newblock \bibinfo{journal}{\emph{Proceedings of the ACM on Management of Data}}  \bibinfo{volume}{1} (\bibinfo{year}{2023}), \bibinfo{pages}{1 -- 18}.
\newblock


\bibitem[\protect\citeauthoryear{Petroni, Rockt{\"{a}}schel, Riedel, Lewis, Bakhtin, Wu, and Miller}{Petroni et~al\mbox{.}}{2019}]%
        {petroni2019language}
\bibfield{author}{\bibinfo{person}{Fabio Petroni}, \bibinfo{person}{Tim Rockt{\"{a}}schel}, \bibinfo{person}{Sebastian Riedel}, \bibinfo{person}{Patrick S.~H. Lewis}, \bibinfo{person}{Anton Bakhtin}, \bibinfo{person}{Yuxiang Wu}, {and} \bibinfo{person}{Alexander~H. Miller}.} \bibinfo{year}{2019}\natexlab{}.
\newblock \showarticletitle{Language Models as Knowledge Bases?}. In \bibinfo{booktitle}{\emph{Proceedings of the 2019 Conference on Empirical Methods in Natural Language Processing and the 9th International Joint Conference on Natural Language Processing, {EMNLP-IJCNLP}}}, \bibfield{editor}{\bibinfo{person}{Kentaro Inui}, \bibinfo{person}{Jing Jiang}, \bibinfo{person}{Vincent Ng}, {and} \bibinfo{person}{Xiaojun Wan}} (Eds.). \bibinfo{publisher}{Association for Computational Linguistics}, \bibinfo{address}{Hong Kong, China}, \bibinfo{pages}{2463--2473}.
\newblock
\urldef\tempurl%
\url{https://doi.org/10.18653/V1/D19-1250}
\showDOI{\tempurl}


\bibitem[\protect\citeauthoryear{Qiao, Zhang, Fang, Luo, Zhou, Jiang, Lv, and Chen}{Qiao et~al\mbox{.}}{2024}]%
        {qiao2024autoact}
\bibfield{author}{\bibinfo{person}{Shuofei Qiao}, \bibinfo{person}{Ningyu Zhang}, \bibinfo{person}{Runnan Fang}, \bibinfo{person}{Yujie Luo}, \bibinfo{person}{Wangchunshu Zhou}, \bibinfo{person}{Yuchen~Eleanor Jiang}, \bibinfo{person}{Chengfei Lv}, {and} \bibinfo{person}{Huajun Chen}.} \bibinfo{year}{2024}\natexlab{}.
\newblock \bibinfo{title}{Autoact: Automatic agent learning from scratch via self-planning}.
\newblock
\newblock


\bibitem[\protect\citeauthoryear{Silver, Huang, Maddison, Guez, Sifre, van~den Driessche, Schrittwieser, Antonoglou, Panneershelvam, Lanctot, Dieleman, Grewe, Nham, Kalchbrenner, Sutskever, Lillicrap, Leach, Kavukcuoglu, Graepel, and Hassabis}{Silver et~al\mbox{.}}{2016}]%
        {alphago}
\bibfield{author}{\bibinfo{person}{David Silver}, \bibinfo{person}{Aja Huang}, \bibinfo{person}{Chris~J. Maddison}, \bibinfo{person}{Arthur Guez}, \bibinfo{person}{L. Sifre}, \bibinfo{person}{George van~den Driessche}, \bibinfo{person}{Julian Schrittwieser}, \bibinfo{person}{Ioannis Antonoglou}, \bibinfo{person}{Vedavyas Panneershelvam}, \bibinfo{person}{Marc Lanctot}, \bibinfo{person}{Sander Dieleman}, \bibinfo{person}{Dominik Grewe}, \bibinfo{person}{John Nham}, \bibinfo{person}{Nal Kalchbrenner}, \bibinfo{person}{Ilya Sutskever}, \bibinfo{person}{Timothy~P. Lillicrap}, \bibinfo{person}{Madeleine Leach}, \bibinfo{person}{Koray Kavukcuoglu}, \bibinfo{person}{Thore Graepel}, {and} \bibinfo{person}{Demis Hassabis}.} \bibinfo{year}{2016}\natexlab{}.
\newblock \showarticletitle{Mastering the game of Go with deep neural networks and tree search}.
\newblock \bibinfo{journal}{\emph{Nature}}  \bibinfo{volume}{529} (\bibinfo{year}{2016}), \bibinfo{pages}{484--489}.
\newblock


\bibitem[\protect\citeauthoryear{Topsakal and Harper}{Topsakal and Harper}{2024}]%
        {topsakal2024benchmarking}
\bibfield{author}{\bibinfo{person}{Oguzhan Topsakal} {and} \bibinfo{person}{Jackson~B Harper}.} \bibinfo{year}{2024}\natexlab{}.
\newblock \showarticletitle{Benchmarking Large Language Model (LLM) Performance for Game Playing via Tic-Tac-Toe}.
\newblock \bibinfo{journal}{\emph{Electronics}} \bibinfo{volume}{13}, \bibinfo{number}{8} (\bibinfo{year}{2024}), \bibinfo{pages}{1532}.
\newblock


\bibitem[\protect\citeauthoryear{Toshniwal, Wiseman, Livescu, and Gimpel}{Toshniwal et~al\mbox{.}}{2021}]%
        {Toshniwal2021ChessAA}
\bibfield{author}{\bibinfo{person}{Shubham Toshniwal}, \bibinfo{person}{Sam Wiseman}, \bibinfo{person}{Karen Livescu}, {and} \bibinfo{person}{Kevin Gimpel}.} \bibinfo{year}{2021}\natexlab{}.
\newblock \showarticletitle{Chess as a Testbed for Language Model State Tracking}. In \bibinfo{booktitle}{\emph{AAAI Conference on Artificial Intelligence}}. \bibinfo{publisher}{{AAAI} Press}, \bibinfo{address}{Virtual}, \bibinfo{pages}{11385--11393}.
\newblock


\bibitem[\protect\citeauthoryear{Yang, Gaglione, Neary, and Topcu}{Yang et~al\mbox{.}}{2022}]%
        {Yang2022AutomatonBasedRO}
\bibfield{author}{\bibinfo{person}{Yunhao Yang}, \bibinfo{person}{Jean-Raphael Gaglione}, \bibinfo{person}{Cyrus Neary}, {and} \bibinfo{person}{Ufuk Topcu}.} \bibinfo{year}{2022}\natexlab{}.
\newblock \bibinfo{title}{Automaton-Based Representations of Task Knowledge from Generative Language Models}.
\newblock
\newblock


\bibitem[\protect\citeauthoryear{Yang and Tomar}{Yang and Tomar}{2023}]%
        {Yang2023OnTP}
\bibfield{author}{\bibinfo{person}{Yunhao Yang} {and} \bibinfo{person}{Anshul Tomar}.} \bibinfo{year}{2023}\natexlab{}.
\newblock \bibinfo{title}{On the Planning, Search, and Memorization Capabilities of Large Language Models}.
\newblock
\newblock


\bibitem[\protect\citeauthoryear{Yao, Yu, Zhao, Shafran, Griffiths, Cao, and Narasimhan}{Yao et~al\mbox{.}}{2024}]%
        {yao2024tree}
\bibfield{author}{\bibinfo{person}{Shunyu Yao}, \bibinfo{person}{Dian Yu}, \bibinfo{person}{Jeffrey Zhao}, \bibinfo{person}{Izhak Shafran}, \bibinfo{person}{Tom Griffiths}, \bibinfo{person}{Yuan Cao}, {and} \bibinfo{person}{Karthik Narasimhan}.} \bibinfo{year}{2024}\natexlab{}.
\newblock \showarticletitle{Tree of thoughts: Deliberate problem solving with large language models}.
\newblock \bibinfo{journal}{\emph{Advances in Neural Information Processing Systems}}  \bibinfo{volume}{36} (\bibinfo{year}{2024}).
\newblock


\end{thebibliography}

%%%%%%%%%%%%%%%%%%%%%%%%%%%%%%%%%%%%%%%%%%%%%%%%%%%%%%%%%%%%%%%%%%%%%%%%

\end{document}